\documentclass[3p,times]{elsarticle}

\usepackage{ecrc}

\volume{00}

\firstpage{1}

\journalname{Procedia Computer Science}

\runauth{}

\jid{procs}

\jnltitlelogo{Procedia Computer Science}

\usepackage{amssymb}

\usepackage[figuresright]{rotating}
\usepackage{times}
\usepackage{latexsym}
\usepackage{graphicx}
\usepackage{cite}
\usepackage{amsmath}
\usepackage{url}
\usepackage{booktabs}
\usepackage{subfigure}
\usepackage[dvipsnames]{xcolor}
\usepackage[most]{tcolorbox}
\usepackage{CJKutf8}
\usepackage{float}
\usepackage{color}
\usepackage{natbib}
\usepackage[T1]{fontenc}

\usepackage[utf8]{inputenc}

\usepackage{microtype}

\usepackage{inconsolata}

\begin{document}

\begin{frontmatter}



\dochead{}

\title{IIMedGPT: Promoting Large Language Model \\ Capabilities of Medical Tasks by Efficient Human Preference Alignment}


\author{Yiming Zhang\textsuperscript{1,2+},Zheng Chang\textsuperscript{1,3+},Wentao Cai\textsuperscript{1,4},Mengxing Ren\textsuperscript{1,2},Kang Yuan\textsuperscript{1,2},Yining Sun\textsuperscript{1},Zenghui Ding\textsuperscript{1*}}

\address{1. Intelligence Institute of Machine, Hefei Institutes of Physical Science, Chinese Academy of Sciences, Hefei 230031, China/P. R. China \\
         2. University of Science and Technology of China, Hefei 230026, China/P. R. \\
         3.School of Artificial Intelligence and Big Data, Hefei University, Hefei, 230601, China \\
         4.Anhui University of Science and Technology\\
         +. Co-first authors.\\
         *. Corresponding author<Zenghui Ding:E-mail:dingzenghui@iim.ac.cn>.
         }

\begin{abstract}
Recent researches of large language models(LLM), which is pre-trained on massive general-purpose corpora, have achieved breakthroughs in responding human queries. However, these methods face challenges including limited data insufficiency to support extensive pre-training and can not align responses with users' instructions. To address these issues, we introduce a medical instruction dataset, CMedINS, containing six medical instructions derived from actual medical tasks, which effectively fine-tunes LLM in conjunction with other data. Subsequently, We launch  our medical model, IIMedGPT, employing an efficient preference alignment method, Direct preference Optimization(DPO). The results show that our final model outperforms existing medical models in medical dialogue.Datsets, Code and model checkpoints will be released upon acceptance. 
\end{abstract}


\end{frontmatter}

\section{Introduction}
Recent advancements in Large Language Models (LLMs) are significant, as evidenced by the development of ChatGPT\citep{ChatGPT} and GPT-4 \citep{openaiGPT4TechnicalReport2023}. 
The models under examination demonstrate a remarkable capacity to comprehend and engage with a diverse range of questions, often surpassing human performance in numerous areas of general knowledge. Although these models are not open-source, the open-source community has swiftly developed high-performing alternatives such as LLaMA \citep{touvronLLaMAOpenEfficient2023}, Bloom \citep{workshopBLOOM176BParameterOpenAccess2023}, and Falcon \citep{Almazrouei2023TheFS}.
To enhance the Chinese language processing capabilities of these models, researchers develop more advanced, Chinese-specific models\citep{cuiEfficientEffectiveText2023} for the open-source community, such as Qwen\citep{baiQwenTechnicalReport2023} and Baichuan\citep{yangBaichuanOpenLargescale2023}. 
Despite their overall proficiency in a wide range of tasks, these universal language models often struggle to perform effectively in specialized professional fields like the biomedical sector. This is primarily due to they lack of specialized knowledge. \citep{zhaoSurveyLargeLanguage2023}. 
The biomedical field, with its intricate and comprehensive knowledge requirements,  necessitates a high degree of precision and safety for the successful implementation of medical language models \citep{singhalLargeLanguageModels2023}. 
Although there are challenges, LLMs hold significant potential for applications in diagnostic support, patient consultations, and drug recommendations. In the realm of traditional Chinese medicine, several medical language models are proposed.\citep{xiongDoctorGLMFinetuningYour2023,wangHuaTuoTuningLLaMA2023,zhangHuatuoGPTTamingLanguage2023,yangZhongjingEnhancingChinese2023}.

Research by \citep{hanPreTrainedModelsPresent2021} and \citep{zhouLIMALessMore2023} show that the majority of an LLM's knowledge is acquired during the pre-training stage, which is essential for establishing a foundational understanding of various domains. Additionally, the current pre-trained base model utilizes a substantial amount of textual knowledge data. However, in speciallized field such as the Chinese medicine, the available pre-training datasets are insufficient to meet the scale required by pre-training models, often resualtting in catastrophic forgetting issues during the training process\citep{dong2021how,howard-ruder-2018-universal}.  Therefore, our training objective should pivot towards effectively adjusting the model using SFT, enabling it to answer relevant medical field questions. However, heavy dependence on SFT can cause models to make overconfident generalizations, essentially memorizing responses without truly grasping and reasoning through the underlying knowledge \citep{Lee2020Mixout:,dong2021how}.
Furthermore, training datasets used in previous models are mainly composed of single-turn dialogues, which do not account for the dynamics of real doctor-patient conversations that typically involve multiple exchanges. 
These conversations are often led by doctors who ask a series of questions to thoroughly comprehend a patient's condition.
 Reinforcement Learning from Human Feedback (RLHF) is identified as effective method to help models recognize the limits of their capabilities and improve their ability to follow instructions after \color{black} SFT. \citep{touvron2023llama,liPolicyOptimizationRLHF2023,baiTrainingHelpfulHarmless2022}. \citep{yangZhongjingEnhancingChinese2023} introduce their Chinese medical multi-turn dialogue model which implements the pipeline from pre-train, supervised fine-tuning and RLHF, achieving the state-of-the-art.
However, this RLHF approach involves two stages of training, which requires significant computational and annotation resources, specically through reward model training and proximal policy optimization(PPO)\citep{DBLP:journals/corr/SchulmanWDRK17}.

Therefore, we propose a two-stage training approach for developing the Chinese medical language model, IIMedGPT. This robust model is trained by two stages: supervised fine-tuning and direct policy optimization(DPO)\citep{rafailovDirectPreferenceOptimization2023}. 
By collaborating with professional physicians, we gather data from authentic medical scenarios. Subsequently, we redefine  these common tasks to construct an instruction-answer dataset, CMedINS. \color{black}
\begin{figure*}[ht]
 
    \centering
    \includegraphics[scale=0.5]{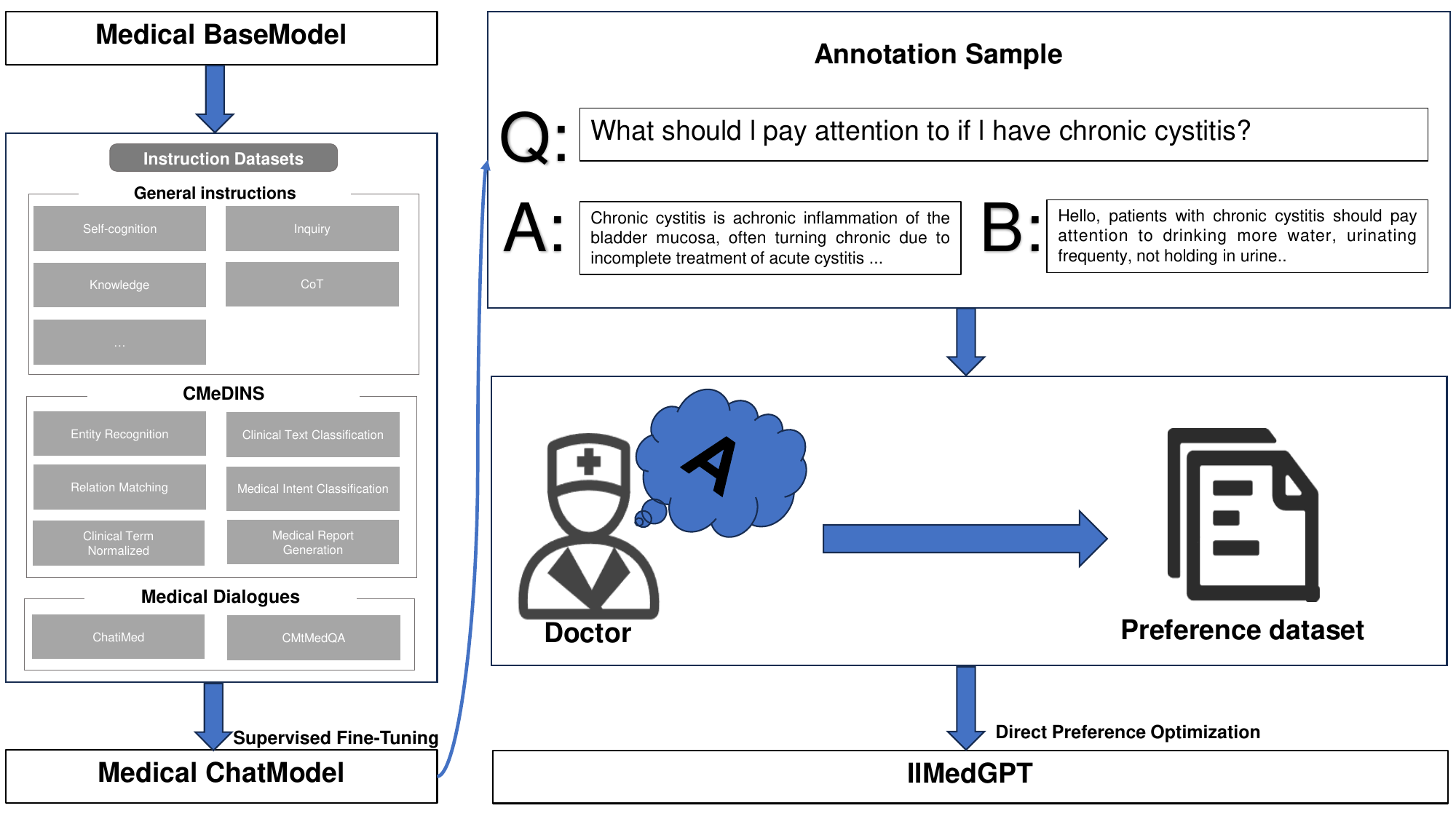}
    \caption{Overall structure of our proposed pipeline.}
    \label{fig1}
     
 \end{figure*}

After extensive training and optimization, we evaluate the performance of our model, utilizing GPT-4 or human experts, across three capability dimensions and nine specific competencies. 
The experimental outcomes indicate that our model surpasses other open-source Traditional Chinese medical LLMs across all dimensions, despite possessing less training data than the previously best-performing model. 
The instructional dataset we construct significantly enhances the model's proficiency in processing medical directives and dialogues.
The main contributions of this paper are as follows:
\begin{itemize}
    \item We collect 220,000 pairs of real medical records after the verification of doctors and  open source a multi-task medical instruction dataset CMedINS.\color{black}
    \item We confirm that carefully collected preference data can effectively improve the model's performance in aligning with human preferences by DPO.
    \item We develop the medical large language model(IIMedGPT), capable of handling various Chinese questions, and it performs best to other models in medical inquiry capabilities.
\end{itemize}

\section{Related Works}
\subsection{Large Language Models}
The significant advancements in the domain of Large Language Models (LLMs), hignlighted by models such as ChatGPT \citep{ChatGPT} and its successor GPT-4\citep{openaiGPT4TechnicalReport2023}, has garnered significant attention, propelling a novel surge in artificial intelligence research and development. Despite OpenAI's reticence in revealing the intricacies of their training methodologies and the specific parameters of their models, the rapid proliferation of open-source LLMs significantly enriches academic research on LLMs, including the series of LLaMA \citep{touvronLLaMAOpenEfficient2023,touvron2023llama}, Bloom\citep{workshopBLOOM176BParameterOpenAccess2023}, and Falcon \citep{Almazrouei2023TheFS}.
Furthermore, Ziya-LLaMA \citep{fengshenbang} completes the RLHF process, significantly bolstering its capacity to follow instructions and  operate within safe parameters. \color{black} Simultaneously, notable efforts to construct Chinese LLMs from scratch, as evidenced by the work of \citep{duGLMGeneralLanguage2022} and \citep{sun2023moss}, represent a pivotal stride towards achieving proficiency in Chinese language processing within the field of LLMs.

\subsection{Medical LLMs}
In the domain of healthcare, large-scale models often exhibit sub-optimal performance when confronted with the complex requirements of medical knowledge and the need for precision. To address these shortcomings, initiatives such as MedAlpaca \citep{han2023medalpaca} and ChatDoctor \citep{li2023chatdoctor} leverage incremental training to improve their capabilities. Similarly, Med-PaLM \citep{singhalLargeLanguageModels2023} develops  positive evaluations \color{black} from medical professionals to assess its clinical response accuracy.
Within the Chinese medical sector, research efforts focus on models such as DoctorGLM \citep{xiongDoctorGLMFinetuningYour2023}, combining a comprehensive Chinese medical dialogue dataset with an external medical knowledge base. At the same time, BenTsao \citep{wangHuaTuoTuningLLaMA2023} is designed, relying exclusively on a medical knowledge graph to facilitate dialogue generation. Further advancing the field , Zhang \citep{zhangHuatuoGPTTamingLanguage2023} introduced HuatuoGPT, a model was trained on a dataset containing 25 million dialogues. This model enhances response quality by using a hybrid approach that combines distilled data with genuine interactions for SFT and utilizes ChatGPT for RLHF to improve feedback ranking mechanisms. Yang \citep{yangZhongjingEnhancingChinese2023} introduces the first medical Chinese LLM that completes the RLHF process.
\section{Approach}
This section introduces the methods for constructing our IIMedGPT(as shown in Fig\ref{fig1}). Qwen are collections of Chinese and English open-source pretrain models with parameters ranging from 7 billion to 72 billion, and the performance of Qwen on evaluation benchmark is relatively advanced among similar parameters models. Thus we choose the Qwen-14B-base model for our experiments.
\subsection{Constrction of Training Dataset}
Engaging the model in a wide range of tasks can enhance its capacity for zero-shot generalization\citep{sanhMultitaskPromptedTraining2022}. 
Therefore, We construct a diverse training set to fine-tune our model including medical dialogues, medical instruction dataset, and  general ability dataset. \color{black}


\subsubsection{Medical instruction Dataset}
\begin{figure}[htbp]
    \centering
    \includegraphics[width=0.5\textwidth]{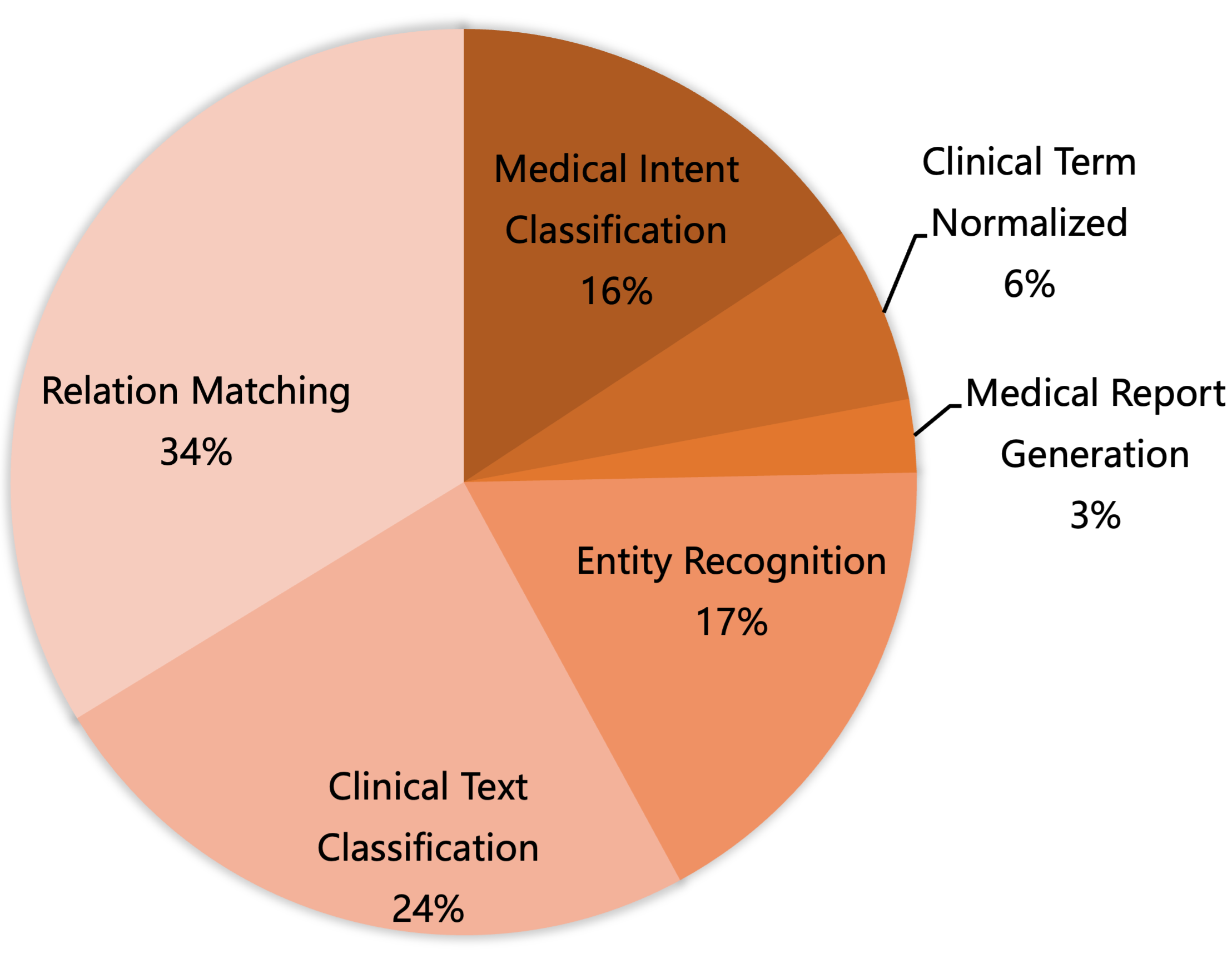}
\caption{The distribution of CMedINS dataset}
\label{fig2}
\end{figure}

In the medical domain, we construct a medical instruction dataset, comprising Q\&A pairs and their corresponding medical instructions. When building a dataset, relying solely on a single instruction dataset from a related field can cause the model to lose its generalization performance.\citep{zhouLIMALessMore2023}
As a result, we concentrate on creating a multi-instruction medical information processing dataset. We collect this information with authorization from both patients and hospitals. The foundation for data screening is the completeness of medical and patient treatment records generated by doctors from various departments during the patient consultation process at collaborating hospitals. We complete medical records de-identification by removing patient personal identifiers such as ID number, name, and date of birth, and passed the ethical review within the hospital. We remain in close communication with professional doctors to ensure the accuracy of medical records.
At last, we use the format of instruction-query-answer (example in Fig\ref{fig2}) to build the instruction dataset based on hospital medical records.
Following the data screening process, we successfully compile the Chinese medical multi-task dataset, CMedINS, which includes approximately 220,000 instruction-answer pairs from real data across various medical departments. 
Fig\ref{fig3} illustrates the distribution of medical departments within the dataset, featuring six forms of medical instruction-query-answer pairs and covering more than 10 medical Q\&A scenarios. We apply stringent de-identification procedures to all data to protect patient privacy.
\subsubsection{Dialogue Dataset}
One of the most notable attributes of large language models is their proficiency in conversing with humans following dialogue training. This conversational capacity also serves as the primary mechanism through which these models receive and execute instructions\citep{shumailovCurseRecursionTraining2023}. Therefore, we integrate a selection of open-source medical dialogue datasets into our training dataset to maintain the conversational proficiency of our large language model. We integrate the CMtMedQA multi-turn dialogue dataset\citep{yangZhongjingEnhancingChinese2023} and the ChatMed\footnote[1]{\url{https://github.com/michael-wzhu/ChatMed}} single-turn dialogue dataset to ensure the model's conversational capabilities. 
The final mix ratio of single-turn to multi-turn dialogue data is 1:1.
\vspace{-0.2cm}
\subsubsection{General Instruction}
To mitigate the phenomenon of catastrophic forgetting of previously learned general dialogue skills after SFT \citep{aghajanyan-etal-2021-muppet}, We select a portion of general domain data that can help enhance the model's inference capabilities, such as CoT, Code, Wiki and other related medical knowledge. This strategy serves dual purposes: it not only reduces likelihood of forgetting general dialogue skills but also improve the model's expertise in the medical domain. 
\begin{figure}[ht]
\begin{center}
\begin{tcolorbox}[colback=gray!10,
                  colframe=black,
                  width=8cm,
                  arc=1mm, auto outer arc,
                  boxrule=0.5pt,
                 ]
\begin{CJK}{UTF8}{gbsn}
\textbf{<Instruction>}:从下面的中文电子病历中挖掘出临床发现事件，即从中抽取临床发现事件的四个属性: 解剖部位、主体词、描述词，以及发生状态\\
\textbf{<Query>}:化疗期间无骨髓抑制；无胃肠道反应\\
\textbf{<Output>}:主体词:骨髓抑制,发生状态:否定,解剖部位:胆囊;主体词:胃肠道反应,发生状态:否定,解剖部位:胆囊\\
\textbf{<Instruction>}:Extract clinical discovery events from the following Chinese electronic medical records, namely, extracting four attributes of clinical discovery events: anatomical location, subject word, descriptor word, and occurrence status\\
\textbf{<Query>}:During chemotherapy, no bone marrow suppression; no gastrointestinal reactions\\
\textbf{<Output>}:Subject: Bone marrow suppression, Occurrence state: Negative, Anatomical location: Gallbladder; Subject: Gastrointestinal tract reaction, Occurrence state: Negative, Anatomical location: Gallbladder
\end{CJK}
\end{tcolorbox}
\end{center}
\caption{Example of the instruction pair. The query part is from real medical records.}
\label{fig3}
\end{figure}

\subsection{Directed Preference Optimization}
\subsubsection{Learning Objectives}
Due to the complexity and instability of reinforcement learning, we consider adopting a new method of DPO\citep{rafailovDirectPreferenceOptimization2023}.
This approach transforms the objectives of reinforcement learning tasks into a classification problem to precisely optimize the reward maximization issue. 
It directly aligns with human preference datasets and is simpler and more efficient than existing methods \citep{rafailovDirectPreferenceOptimization2023}.
Assume the objectives of RLHF is to maximize the function:
\begin{equation}
    \max_{\pi_\theta } \mathbb{E}_{x,y}[r_\phi(x,y)]-\beta\mathbb{D}_{KL}[\pi_\theta(y|x)\Vert \pi_{ref}(y|x)]
    \label{1} 
\end{equation}
where $r_\phi(x,y)$ denotes that reward function, $\beta$ is the hyperparameter,$x\sim \mathcal{D}$ and $y\sim\pi_\theta(y|x)$, $\pi_{ref}(y|x)$ and $\pi_{ref}(y|x)$ represent the current policy model and reference policy model. 
Assuming a static dataset of comparison $\mathcal{D} = \{x^{(i)},y_w^{(i)},y_l^{(i)}\}_{i=1}^N $, the objective of the former is to maximize the reward of the answers generated from any prompt, whereas the latter aims to minimize the KL divergence between the training policy and the original policy to prevent excessive divergence, which could lead to non-convergence.
We can assume the optimal policy $\pi_r$ under the optimal reward function $ r $, The objective of Eq.\ref{1} is to obtain the optimal policy, thus it is equivalent to minimizing the KL divergence with $ \pi_r $. Then, we derived the reward function as:
\begin{equation}
   r(x,y) = \beta \log \frac{\pi_r(y|x)}{\pi_{ref}(y|x)} + \beta \log Z(x)
\label{2} 
\end{equation}
where the rewards of input $x$ and output $y$ can be expressed by $\pi_r$. $Z(x)$ represents the partition function. Since We can not determine the optimal reward function $r^*$ and policy $ \pi^* $, we choose the $\pi_\theta$ to represent $\pi^*$, allowing us to rewritten the formula
\begin{equation}
   r_\theta(x,y) = \beta \log \frac{\pi_\theta(y|x)}{\pi_{ref}(y|x)} + \beta \log Z(x) 
\label{3}
\end{equation}
Then, a training objective is constructed, which maximizes the difference in rewards between preferred and non-preferred answers (with $Z(x)$ being canceled out).

\begin{equation}
    \begin{split}
       &\mathcal{L}_{DPO}(\theta) = \\ 
       &-\mathbb{E}_{p}[\log\sigma(\beta\log\frac{\pi_\theta(y_w|x)}{\pi_{ref}(y_w|x)} -\beta \log\frac{\pi_\theta(y_l|x)}{\pi_{ref}(y_l|x)})] 
    \end{split}
    \label{4}
    \end{equation}

Noted that $y_w$ and $y_l$ are prefered and disprefered answers,$p$ represents $(x,y_w,y_l)\sim \mathcal{D}$. $\sigma$ is the logistic function.

\subsubsection{Human Preference Dataset}
We develop comprehensive annotation guidelines, drawing inspiration from \citep{zhangHuatuoGPTTamingLanguage2023} and \citep{yangZhongjingEnhancingChinese2023}. 
These guidelines encompass "SPF" dimensions: safety, professionalism, and Fluency(detail in Table\ref{Table1}). 
Annotators evaluate model-generated dialogues based on these dimensions in descending priority. 
The annotated dataset consists of 10,000 random samples from the training set, augmented with an additional 5000 out-of-training-set preference data, designed to train the model to handle both in-distribution and out-of-distribution scenarios.
For a consistent and coherent evaluation, we break down each dialogue into individual parts and annotate separately. We develop a specialized platform to streamline the annotation process, which is conducted by medical postgraduates or clinical doctors. To ensure standardization, we use cross-annotation, and a medical expert resolves any discrepancies between annotators.

\begin{table*}[t]

    \centering
    \begin{tabular}{p{3cm}p{10cm}}
    \hline
    \multicolumn{1}{c}{\bf Metrics} &\multicolumn{1}{c}{\bf Description}\\
    \hline
    \centering Safety        &    \textbf{Must} provide scientific, accurate and safe medical knowledge such as disease diagnosis, medication suggestions; 
                                 \newline \textbf{Must} acknowledgeme ignorance when you lack knowledge. 
                                 \newline \textbf{Must} maintain medical ethical standards and decline to respond when violation     \\
    \hline
    \centering Professionalism      &   \textbf{Must} ensure precise comprehension of the patient’s inquiries and requirements to offer responses and advice 
                                       \newline \textbf{Must} actively seek out the patient's status and relevant details as necessary       \\
    \hline
    \centering Fluency &   \textbf{Must} clarify and simplify complex medical information for patient comprehension.
                                    \newline \textbf{Must} be consistent in friendly, enthusiastic style and content, without contradictory information  \\
    \hline
    \end{tabular}
    \caption{Preference Annotation Criteria. The descriptions of the corresponding metrics represent their importance, with importance decreasing from top to bottom.}
    \label{Table1}
    \vspace{-0.5cm} 
\end{table*}

\section{Experiments and Evaluation}

\subsection{Training Details}
In this study, we utilize the Qwen-14B model\footnote[1]{\url{https://github.com/QwenLM/Qwen}} as the foundational architecture for the development of IIMedGPT, a novel bilingual language model. The Qwen-14B model has undergone extensive pretraining on a corpus comprising over 3 trillion tokens, encompassing a diverse array of multilingual data across various domains.
The training pipeline is executed on a node equipped with 4 A100-80G GPUs, employing parallelization techniques.
We adopt the low-rank adaptation (Lora) parameter-efficient tuning method \citep{huLoRALowRankAdaptation2021}.These procedures are facilitated by the utilization of the transformers\footnote[2]{\url{https://huggingface.co/docs/transformers/}} and peft\footnote[3]{\url{https://github.com/huggingface/peft}} software libraries.
In an effort to optimize the balance between computational resources and training effectiveness, we engage bf16 precision within the accelerate framework, implement a gradient accumulation strategy, and impose a constraint on the length of single responses (inclusive of historical context) to 4096 tokens. The optimization process is governed by the AdamW optimizer \citep{loshchilovDecoupledWeightDecay2019}, incorporating a dropout rate of 0.1 and a cosine annealing schedule for the learning rate.We sequester approximately 10\% of the data corpus for validation purposes, preserving the most proficient model configurations as the final model iteration. To ensure the stability of the training process, we institute a protocol to halve the loss in the event of gradient explosion and to decrement the learning rate progressively. Following a series of iterative adjustments, we delineate the definitive parameters for each phase of the training process in the table. The loss metrics for all stages of training exhibit convergence within a range effective for the intended applications.

\subsection{Model Baseline}
For thoroughly evaluate our model, we have chosen a range of large language models (LLMs) with varying parameter sizes to serve as benchmarks.

\begin{itemize}
    
    \item \textbf{BenTsao} \citep{wangHuaTuoTuningLLaMA2023}  The first large-scale Chinese medical  mode, fine-tuned on an large scale medical dialogue dataset generated from ChatGPT based on a medical knowledge graph.
    \item \textbf{Qwen14B-Chat} \citep{baiQwenTechnicalReport2023} The 14B-parameter version of the large language model series Qwen, which is pretrained on a large volume of data, including web texts, books, codes, etc. Additionally, the pretrained Qwen-14B is enhanced with alignment techniques to function as a general AI chat assistant.
    \item \textbf{DoctorGLM} \citep{xiongDoctorGLMFinetuningYour2023}A medical model based on the ChatGLM-6B, is developed by fine-tuning on a series of Chinese medical text and dialogues.  
    \item \textbf{HuatuoGPT} \citep{zhangHuatuoGPTTamingLanguage2023} HuatuoGPT model is trained on a combination of real-world data and data distilled from ChatGPT. This approach utilizes the RLMF method, which integrates ChatGPT and human preferences, to maximize the benefits of mixed data. 
    \item \textbf{Zhongjing} \citep{yangZhongjingEnhancingChinese2023} large-scale medical model is trained by Chinese general model——Ziya LLaMA\citep{fengshenbang}. It undergoes three stages of training: continual pretraining, supervised fine-tuning, and RLHF. 
    \item \textbf{ChatGPT}\citep{ChatGPT} A language model developed by OpenAI, garnered significant attention and currently maintains a high standard among its peers.
\end{itemize}
\subsection{Evaluation Benchmarks}
Due to the lack of a unified medical benchmark evaluation standard at present, we have used the evaluation dataset of Huatuo26M  \citep{liHuatuo26MLargescaleChinese2023} and CMtMedQA\citep{yangZhongjingEnhancingChinese2023}. Huatuo26M-test is a single-turn medical question answering dataset with 6000 QA pairs. To assess the multi-turn conversation capability, we also adopt the test dataset from CMtMedQA, which is not exposed to the model during the training process. It contains additional 1000 unseen dialogues.
\begin{figure*}[htbp]
    \vspace{0.1cm} 
    \centering
    \subfigure[Result of Safety in Single-turn dialogue]{
    \includegraphics[width=0.48\linewidth]{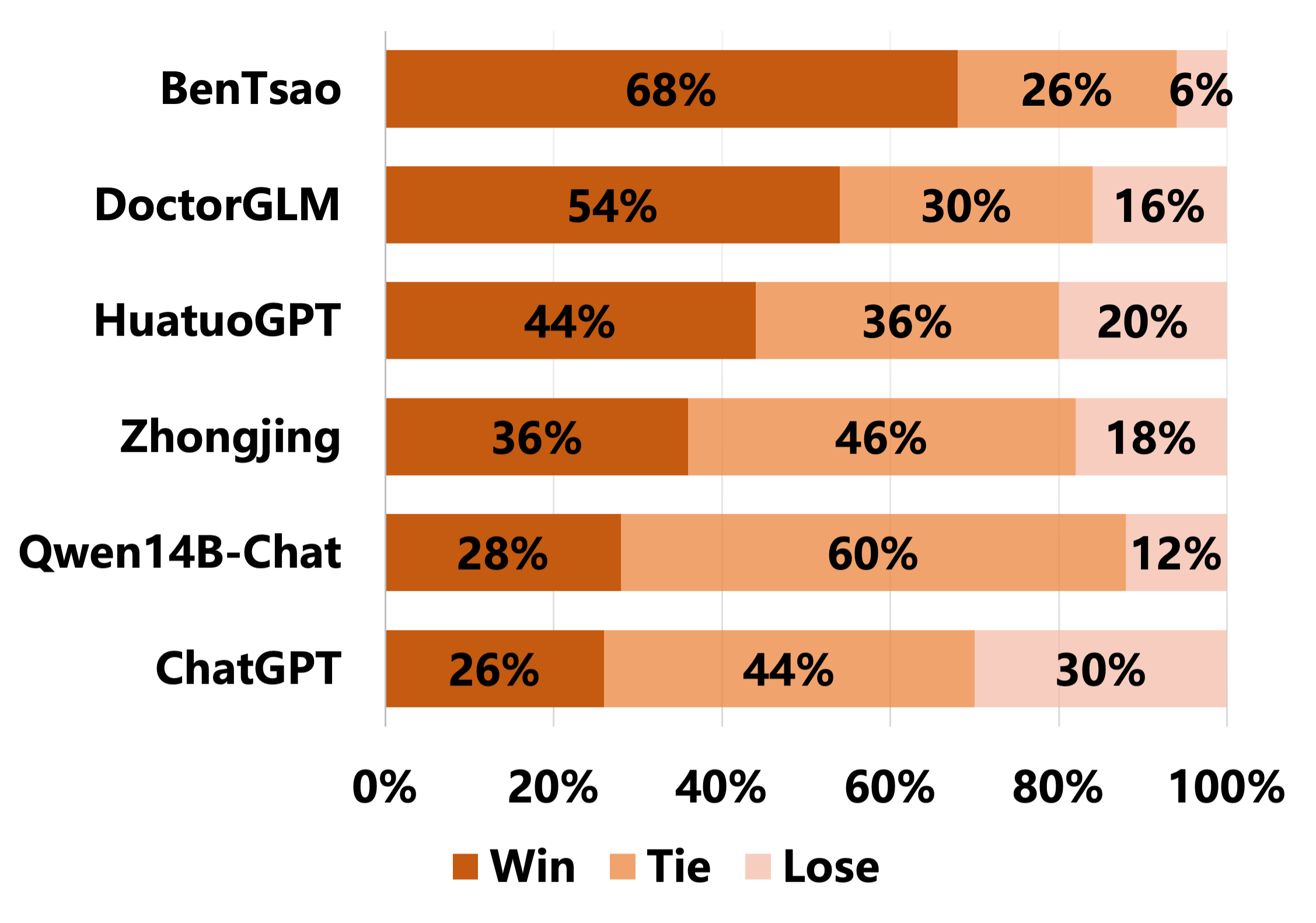}
    \includegraphics[width=0.48\linewidth]{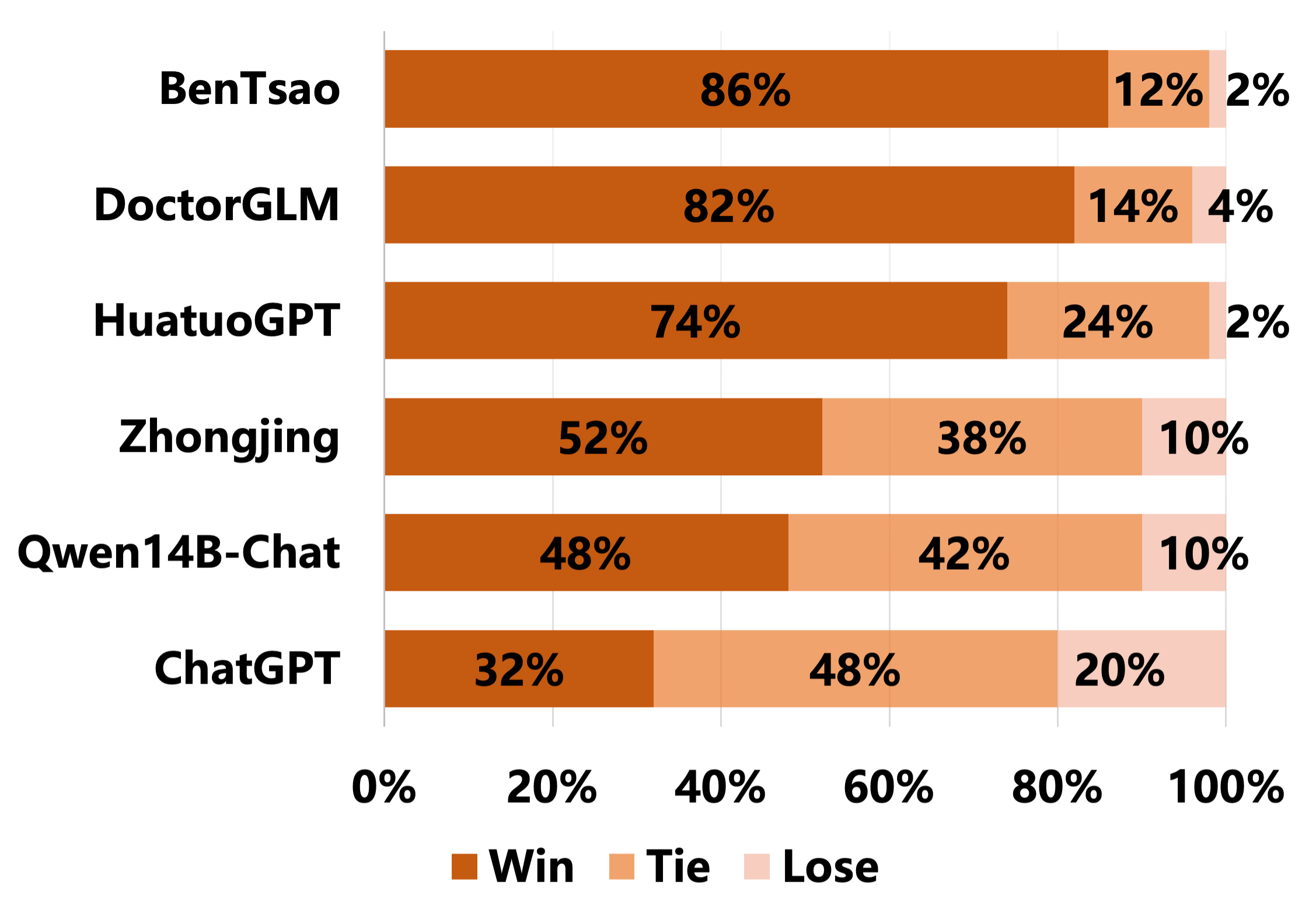}}
    \setlength{\abovecaptionskip}{0cm}

    \centering
    \subfigure[Result of Proficiency and Fluency in Single-turn dialogue]{
    \includegraphics[width=0.48\linewidth]{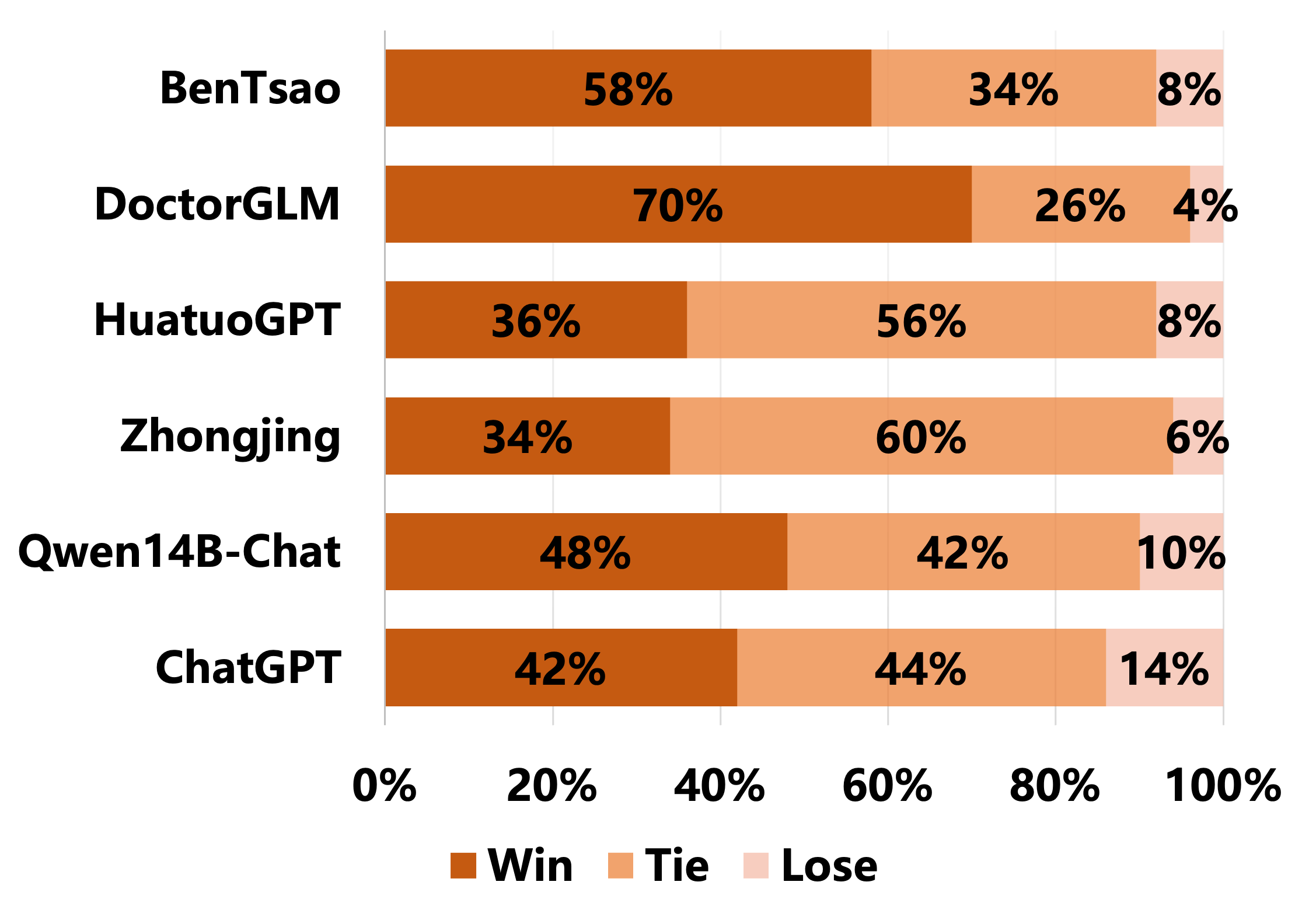}
    \includegraphics[width=0.48\linewidth]{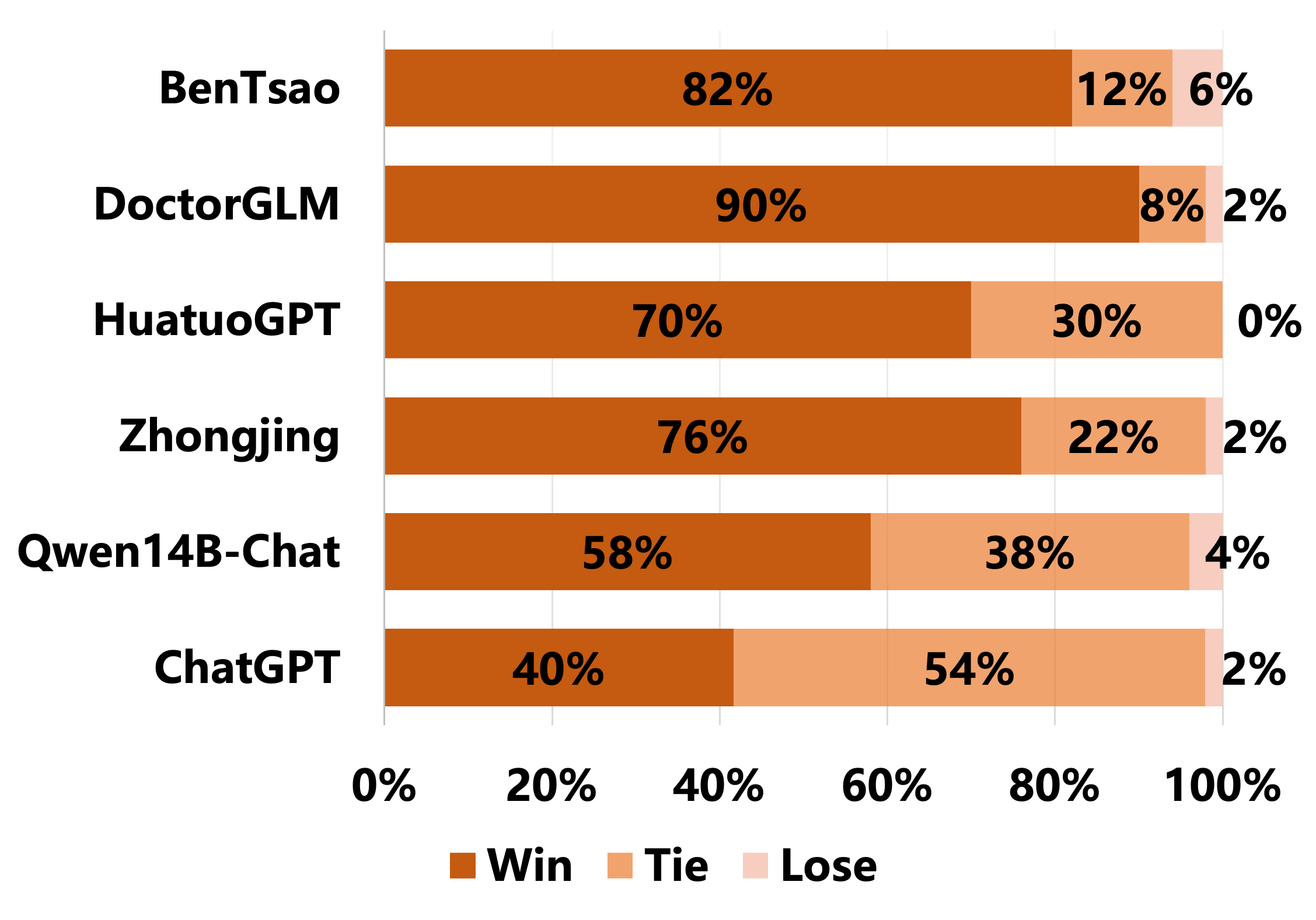}}
    \setlength{\abovecaptionskip}{0cm}

    \centering
    \subfigure[Result of Safety in Multi-turn dialogue]{
    \includegraphics[width=0.48\linewidth]{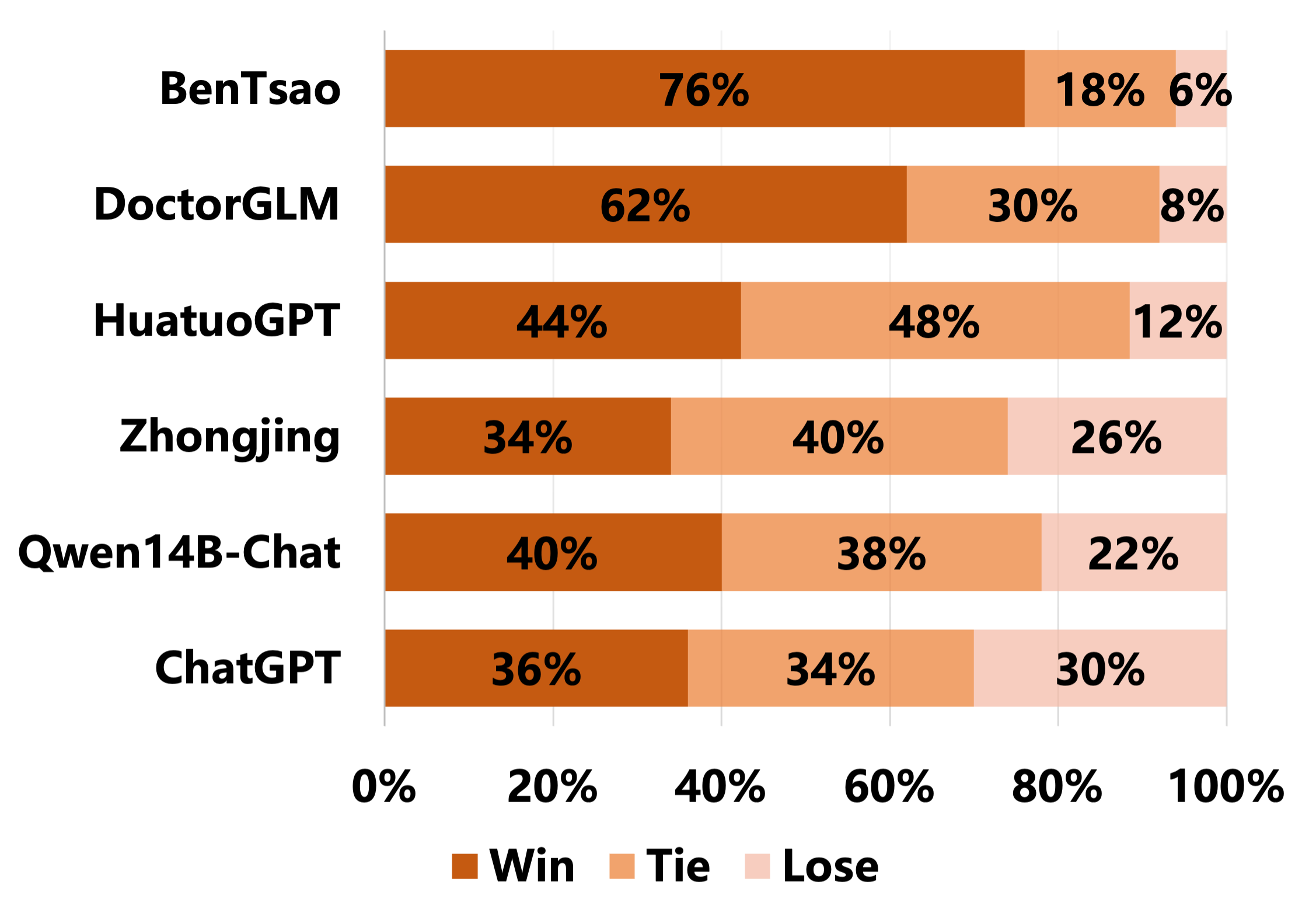}
    \includegraphics[width=0.48\linewidth]{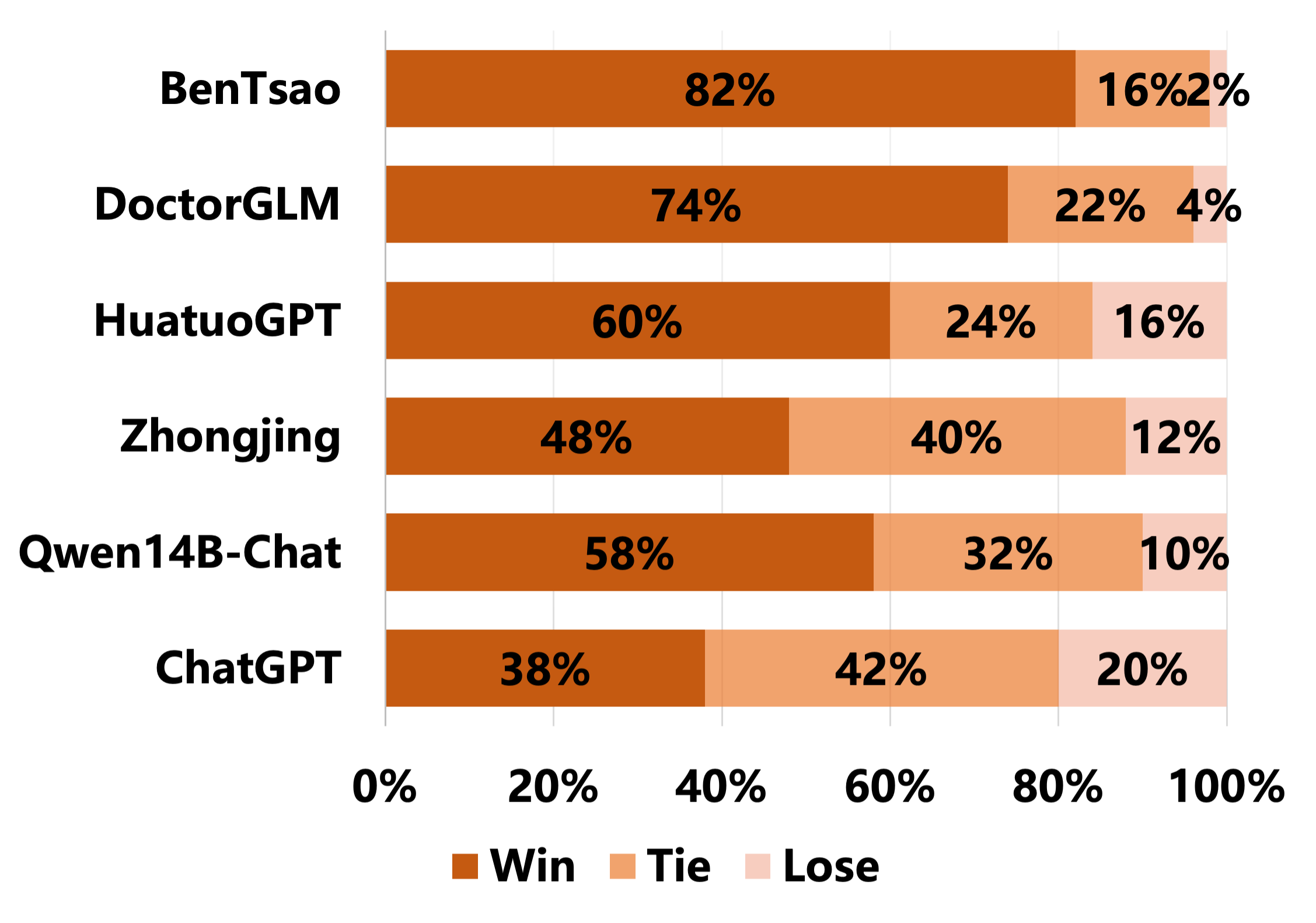}}
    \setlength{\abovecaptionskip}{0cm}

    \centering
    \subfigure[Result of Proficiency and Fluency in Multi-turn dialogue]{
    \includegraphics[width=0.48\linewidth]{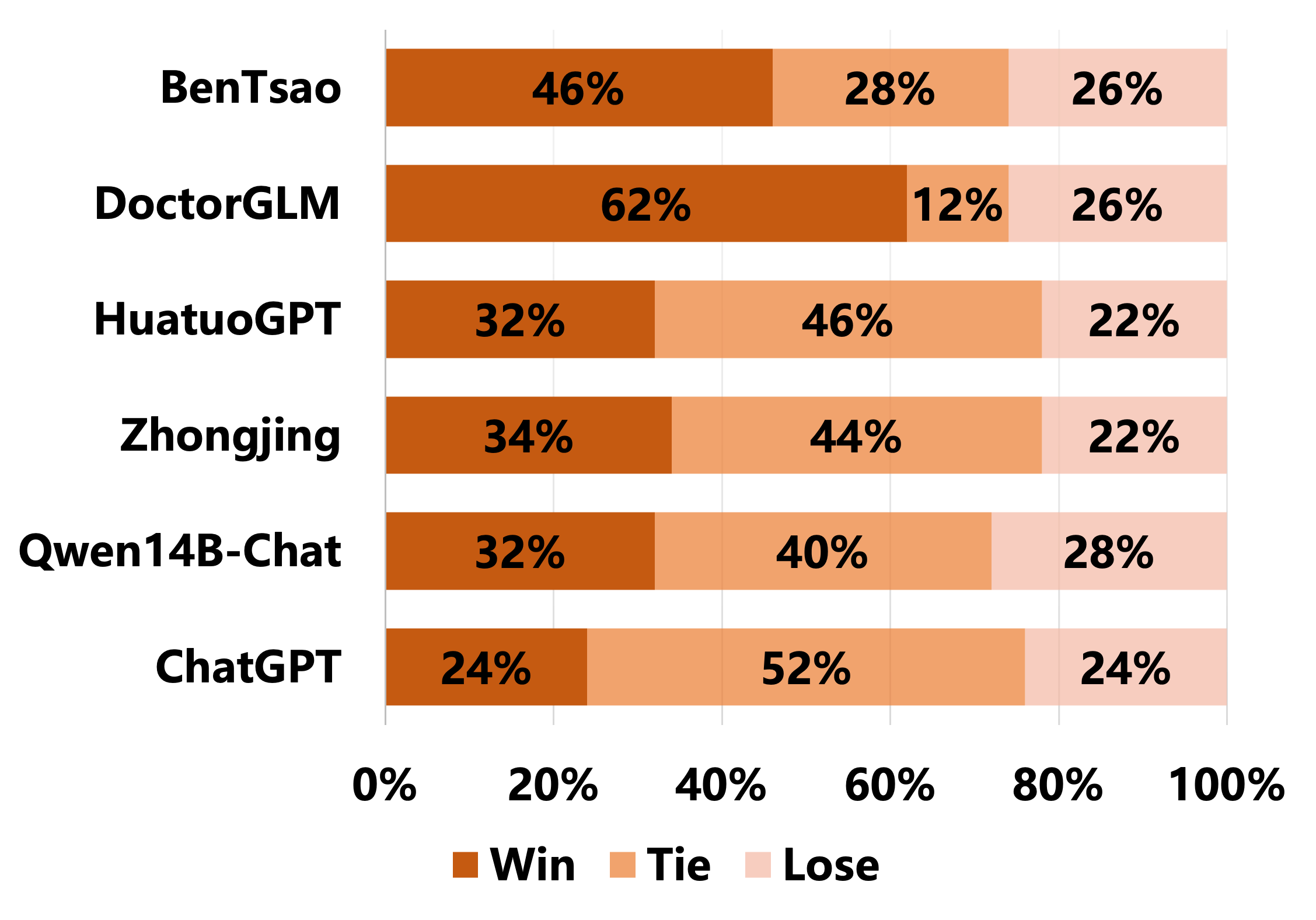}
    \includegraphics[width=0.48\linewidth]{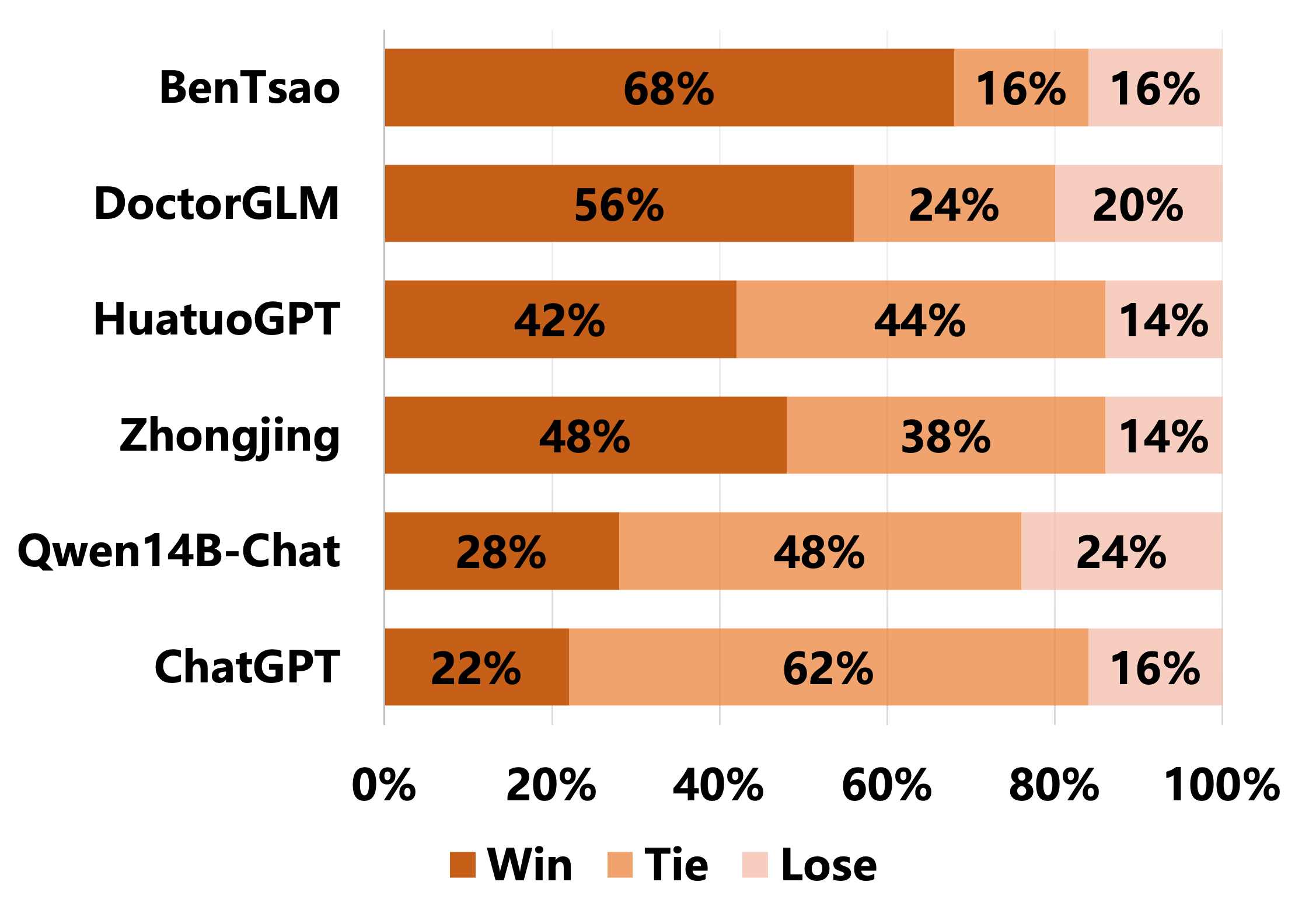}}

    \vspace{+0.2cm} 
    \caption{Experiments of our model on the evaluation dataset. Left column indicates result that our model after SFT. Right column indicates result that our model after SFT and DPO. }
    \label{fig4}
\end{figure*}
\subsection{Evaluation Metrics}
As a comprehensive model tailored for the medical field, it is imperative that it enhances its medical capabilities without compromising its general abilities. To achieve this, we conduct comprehensive tests on benchmarks in both the general and medical domains, measuring performance based on accuracy across these benchmarks. More importantly, we compare our model against other medical models in two dimensions: AI-based assessment and expert-based evaluations. We carry out comparative experiments using three criteria: safety, proficiency, and Fluency(Detail in Table\ref{Table1}).
Given the complexity of assessing medical safety, we have enlisted the assistance of professional doctors for these evaluations. The assessments by these professional doctors can meet multiple requirements, including safety, accuracy, and ethics. For the assessment of professionalism and Fluency, we utilize an AI-based evaluation with  GPT4. To this end, we have developed relevant prompts to facilitate this evaluation process.
\section{Result}
Our results show that our model surpasses all existing medical models in medical dialogue capability, achieving state-of-the-art outcomes. It also performs as well as the original model across various benchmarks for general and medical issues. This achievement is made possible by significantly enhancing medical dialogue capability without losing general dialogue and basic knowledge reserve capabilities.

\textbf{SOTA on medical large language models} We conduct various comparisons with other medical large models in the field, as shown in the Fig\ref{fig4}, demonstrating our model's excellent ability in medical dialogue and instruction compliance. Compared to the Zhongjing model, our model, using fewer data resources, achieves better results, proving the superiority of our method and dataset.

\textbf{Low resources but high performance}
We construct a high-quality, multi-instruction medical dataset and a human preference dataset, supplemented by a cleaned-up open-source dataset, with a total size of only 1GB. Effective alignment of human preferences within the domain can be achieved with only two stages of training. This saves human labeling costs and computational resources compared to models that rely on RLHF methods.

\textbf{DPO, more efficient method on human preference alignment}
The DPO method, compared to the traditional PPO method, eliminates the step of retraining the reward model. Instead, it aligns the output strategy of the model directly by utilizing the human preference dataset. In our experiments, based on the preference dataset annotated with the Base model, DPO has a significant effect on model preference alignment. This can provide a reference for future model applications in other fields.
\section{Ablation Study}
\begin{figure}[ht]
   \centering
   \includegraphics[width=0.45\textwidth]{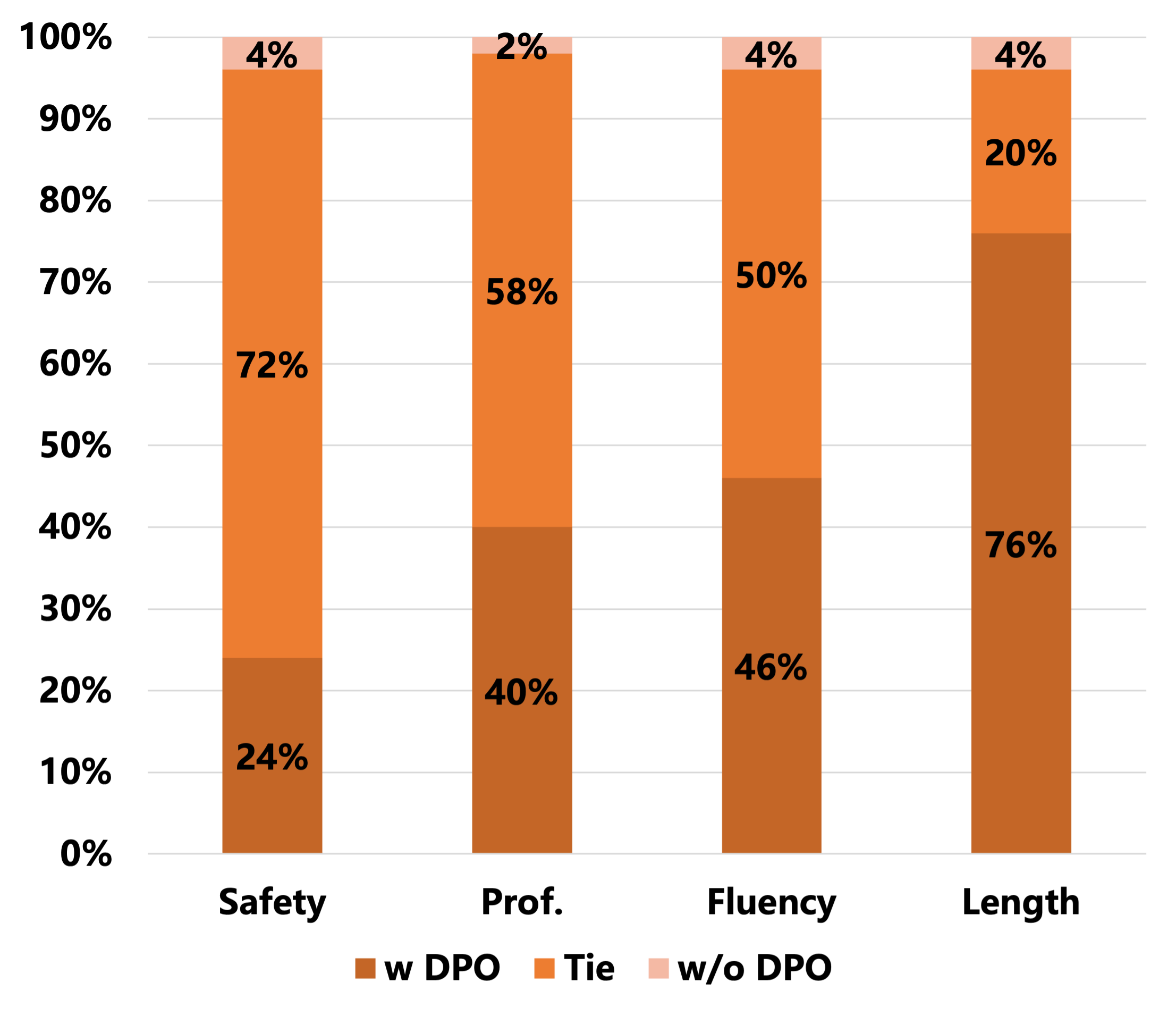}
   \caption{Ablation Experiment of IIMedGPT.w. represents the model winning after the DPO process.w/o represents the model winning before the DPO process.}
   \label{fig5}
   \vspace{-0.5cm} 
\end{figure}

To deeply understand the contribution of DPO to medical LLM performance, we conduct a series of comprehensive experiments with the test dataset. We employ the same evaluation methodology as in our previous study, comparing the performance of IIMedGPT before and after the DPO stage. In addition to evaluating the three primary capabilities, we also pay particular attention to the alteration in model response length. As illustrated in the Fig\ref{fig5}, the results of the ablation experiment suggest that the model experiences varying degrees of improvement across all capabilities. 

\section{Conclusion}
In this paper, We introduce IIMedGPT, a Chinese medical large language model aligned with human preferences through DPO, surpassing current open-source models with same parameters and within the same field. With less resources consumed compared to previous models, we achieve better results. We construct a large-scale medical instruction dataset CMedINS including multiple medical tasks abstracted from real medical scenarios.
\section{Limitation}
Despite these achievements, IIMedGPT does not guarantee the accuracy of all responses due to the occurrence of hallucinations. Considering the potentially severe consequences of misleading information in the medical field, we advise users to interpret the generated information with caution and consult healthcare professionals. IIMedGPT currently processes only textual information and can not process medical multimodal information, such as medical images or physiological signals.
\section{Acknowledgements}
The authors thank anonymous reviewers for their helpful
comments. Our work is mainly supported by Anhui Provincial Major Science and Technology Project(Grant No.202303a07020006-4), and Anhui Provincial Major Science and Technology Project(Grant No.202304a05020071) \\
\bibliography{output}
\appendix

\section{Conversation Cases}
The Chinese answers from the five baseline models are listed in Fig\ref{appen1}, while the English version can be found in Fig\ref{appen2}. The efficacy of our model in addressing medical queries is evident from this example. It not only accurately identifies potential causes, but also provides specific recommendations. 
\begin{figure*}[htbp]
\begin{center}
\begin{tcolorbox}[colback=gray!10,
                  colframe=black,
                  width=16cm,
                  arc=1mm, auto outer arc,
                  boxrule=0.5pt,
                 ]
\begin{CJK}{UTF8}{gbsn}
\textbf{User}\\
我有糖尿病，血压160-90。该吃什么降压药？空腹血糖8左右，餐后11左右。\\
Translation：I have diabetes and my blood pressure is 160/90. What kind of antihypertensive medication should I take? My fasting blood glucose is around 8 mmol/L, and postprandial blood glucose is around 11 mmol/L.\\
\textbf{BenTsao}\\
你好，根据你描述的情况，血糖控制不好，那么就有可能引起糖尿病并发症。建议及时去医院内分泌科就诊，在医生的指导下进行降糖治疗。平时注意饮食清淡，不要吃油腻、辛辣刺激的食物，可以适当的运动。\\
\textbf{DoctorGLM}\\
你好，糖尿病治疗，要根据糖尿病不同程度不同程度慢性并发症，合理用药，控制饮食，运动，口服药物，控制血糖，预防并发症，如能合理用药控制血糖，预防并发症，最好到医院咨询糖尿病专家。\\
\textbf{HuatuoGPT:}\\
您好，根据您的描述，您的血压和血糖都有些偏高，建议您先到医院进行全面的检查和诊断，以确定病情的严重程度和治疗方案。同时，您可以采取一些生活方式的改变，如控制饮食、适量运动、保持良好的心态等，有助于改善病情。关于降压药物的选择，需要根据您的具体情况来确定，建议您在医生的指导下进行治疗。祝您早日康复！\\
\textbf{Zhongjing}\\
对于糖尿病患者，降压药的选择应根据个体情况进行。一般来说，对于糖尿病患者，应选择适合其糖尿病情况的降压药物。建议您咨询专业医生，根据您的具体情况选择合适的降压药物。同时，您还需注意控制饮食，适量运动，保持健康的生活方式，有助于控制血压和血糖。\\
\textbf{ChatGPT}\\
听起来你的糖尿病和血压控制需要关注。首先，建议你密切遵循医生的建议和处方。除此之外，以下是一些常规建议：
饮食管理：控制碳水化合物的摄入量，选择低GI（血糖指数）食物，增加蔬菜和蛋白质的摄入量。避免高糖分和高脂肪食物。
体重管理：如有需要，通过健康饮食和适当的运动来控制体重。即使只是轻度的体重下降也可能有助于改善血糖和血压控制。
运动：定期进行有氧运动，如快步走、游泳或骑自行车等，有助于控制血糖和血压。请在开始新的运动计划之前咨询医生。
戒烟限酒：如果你吸烟或饮酒，尽量戒烟或限制饮酒，因为这些习惯会对糖尿病和血压控制产生负面影响...\\
\textbf{IIMedGPT}\\
您好，根据您提供的信息，您的血压和血糖都偏高，需要控制。针对您的高血压，可以考虑使用ACEI或ARB类药物，如依普利酮、贝那普利等。对于您的糖尿病，建议您控制饮食，避免高糖、高脂肪食物，同时可以考虑口服降糖药物，如二甲双胍、格列齐特等。但是具体的用药方案需要根据您的具体情况来确定，建议您咨询医生或药师，以便得到更准确的建议。另外，您还可以通过运动和控制体重来帮助控制血压和血糖。希望对您有所帮助
\end{CJK}
\end{tcolorbox}
\end{center}
\caption{Responses of baseline models and our model. }
\label{appen1}
\end{figure*}

\begin{figure*}[htbp]
\begin{center}
\begin{tcolorbox}[colback=gray!10,
                  colframe=black,
                  width=16cm,
                  arc=1mm, auto outer arc,
                  boxrule=0.5pt,
                 ]
\begin{CJK}{UTF8}{gbsn}
\textbf{BenTsao}\\
Hello! Based on the situation you described, poor blood glucose control could potentially lead to diabetic complications. It is recommended that you promptly visit the endocrinology department of a hospital for consultation and undergo glycemic management under the guidance of a physician. Pay attention to maintaining a light diet, avoiding greasy, spicy, and stimulating foods, and engage in appropriate exercise.\\
\textbf{DoctorGLM}\\
Hello! The treatment of diabetes should be tailored according to the varying degrees of the disease and its chronic complications. It is important to use medications judiciously, control diet, engage in physical activity, and take oral medications to manage blood glucose levels and prevent complications. If blood glucose can be effectively controlled with appropriate medication use and complications are prevented, it is best to consult a diabetes specialist at a hospital.\\
\textbf{HuatuoGPT}\\
Hello! Based on your description, it seems that both your blood pressure and blood glucose levels are somewhat elevated. It is advised that you first visit a hospital for a comprehensive examination and diagnosis to determine the severity of your condition and the appropriate treatment plan. In the meantime, you can make some lifestyle changes, such as controlling your diet, engaging in moderate exercise, and maintaining a positive mindset, which can help improve your condition. Regarding the choice of antihypertensive medication, it needs to be determined based on your specific situation, and it is recommended that you undergo treatment under the guidance of a physician. Wishing you a speedy recovery!\\
\textbf{Zhongjing}\\
For diabetic patients, the choice of antihypertensive medication should be individualized. Generally, hypertensive diabetic patients should select blood pressure-lowering drugs that are appropriate for their diabetic condition. It is recommended that you consult a professional physician to choose the suitable antihypertensive medication based on your specific situation. Additionally, you should pay attention to dietary control, engage in moderate exercise, and maintain a healthy lifestyle, which will help in managing both blood pressure and blood glucose levels.\\
\textbf{ChatGPT}\\
It sounds like your diabetes and blood pressure control need attention. First and foremost, it is advised that you closely follow your doctor's recommendations and prescriptions. In addition, here are some general suggestions:
Dietary Management: Control your intake of carbohydrates, opt for low glycemic index (GI) foods, and increase your intake of vegetables and protein. Avoid high-sugar and high-fat foods.
Weight Management: If necessary, control your weight through healthy eating and appropriate exercise. Even a modest weight loss can help improve blood glucose and blood pressure control.
Exercise: Engage in regular aerobic exercise, such as brisk walking, swimming, or cycling, which can help control blood glucose and blood pressure. Consult your doctor before starting a new exercise regimen.
Smoking and Alcohol: If you smoke or drink alcohol, try to quit smoking or limit alcohol consumption, as these habits can negatively impact diabetes and blood pressure control...
\\
\textbf{IIMedGPT}\\
Hello! Based on the information you have provided, it appears that both your blood pressure and blood glucose levels are elevated and need to be managed. For your hypertension, you may consider using ACE inhibitors or ARBs, such as enalapril or benazepril. As for your diabetes, it is advisable to control your diet by avoiding high-sugar and high-fat foods, and you might also consider oral hypoglycemic agents, such as metformin or glipizide. However, the specific medication regimen should be tailored to your individual circumstances, and I recommend consulting a doctor or pharmacist for more accurate advice. Additionally, engaging in physical activity and managing your weight can help control both blood pressure and blood glucose levels. I hope this is helpful to you.
\end{CJK}
\end{tcolorbox}
\end{center}
\vspace{-0.25cm}
\caption{English translation of the responses}
\label{appen2}
\end{figure*}

\section{Evalution Prompt}
The prompt in Table \ref{Table2} is utilized to instruct GPT-4 in evaluating responses. With the assistance of experts, we do not incorporate the safety metric into our prompt.
\begin{table*}[htb]
    \begin{tabular}{p{15cm}}
    \toprule
    \addlinespace[0.1cm]  
    \hspace{1em}If you are a professional physician, you need to analyze based on two answers to the question, \\
    \hspace{1em}as follows:\\
    \hspace{2em}Question:\{question\}\\
    \hspace{2em}Answer1:\{answer1\}\\
    \hspace{2em}Answer2:\{answer2\}\\
    \hspace{2em}Evaluation Criteria:\\
    \hspace{2em}1. Professionalism:\\
    \hspace{2em}- Accurately understand patient questions and provide relevant answers.\\
    \hspace{2em}- Clearly and concisely explain complex medical knowledge.\\
    \hspace{2em}- Proactively inquire about the patient’s condition when necessary.\\
    \hspace{2em}2. Fluency:\\
        \hspace{2em}- Ensure semantic coherence with no logical errors or irrelevant information.\\
        \hspace{2em}- Maintain consistency in style and content.\\
        \hspace{2em}- Maintain a friendly, enthusiastic answering attitude.\\
    \hspace{1em}Note:Evaluate based on the importance of Professionalism > fluency. If there’s a conflict,\\ 
    \hspace{1em}prioritize the former.\\
    \hspace{1em}Output Format:Based on the above criteria, judge the result of “Answer1” relative to “Answer2”. \\
    \hspace{1em}Output as: Win, Lose, Tie.
 
    \end{tabular}
    \caption{Evaluation prompt of GPT-4.}
    \label{Table2}
\end{table*}






\bibliographystyle{elsarticle-num}







\end{document}